# Re-expression of manual expertise through semi-automatic control of a teleoperated system

Erwann Landais[1], Nasser Rezzoug [2] and Vincent Padois[1]

[1]INRIA, 33400 Talence, France

[2]Institut Pprime (CNRS, Université de Poitiers, ISAE-ENSMA), UPR 3346, 86360 Poitiers, France

**ABSTRACT**

While the search for new solvents in the chemical industry is of uttermost importance with respect to environmental considerations, this domain remains strongly tied to highly manual and visual inspection of dangerous products by experts. Our proposal to guarantee safe and efficient performance of this task is to deport product manipulation to a robotic arm, over which the user has limited control. This proposal was tested in an experiment in which participants were invited to perform a similar task via direct handling, then under conditions secured by a protection barrier, and finally with teleoperation using different trajectory modulation variants. The data recovered showed that although this task is indeed achievable via the proposed interface, the proposed variants fail to achieve satisfactory performance regarding execution time.

## INTRODUCTION

### Motivation

Before a chemical solution can be marketed, a solvent must be found which complies with environmental and health standards and solubilises the solute. In the chemical industry, this is generally done by visually assessing the solvency of the considered solute in a range of solvents. This requires considerable expertise (Abbott et al., 2008). As the chemicals handled may imply a critical danger (CMR substances, explosion or heat release), mechanical protection barriers are used (fume hoods, gloveboxes). As a result, experts are operating in dangerous and difficult conditions, with considerable physical and cognitive constraints.

Carrying out this task using a remotely controlled robot to reproduce the movements of the vials containing the chemical solutions is a potential solution to alleviate these constraints while retaining the contribution of user visual and cognitive expertise. This solution will only be relevant if it intuitively transcribes the manual expertise of its users (to avoid any degradation of visual and cognitive expertise), and respects the manual requirements of this task (i.e. large-amplitude orientation movements in a space restricted by a protective hood, without unintentionally degrading the visual characteristics of the object).

Generating a priori unknown sequences of movements online with a robot in a limited workspace is not without risk, especially when performance is at stake. Our proposal is to separate control of the movement into two parts: control of the path (set of spatial poses) and of the trajectories associated with this path (velocity along the path). The user can then partially control the robot's movements, by choosing the type of generic path, and modulating the trajectory performed on this path in real time through an Online Trajectory Generator such as the one proposed in (Berscheid & Kröger, 2021). In order to verify whether this type of teleoperated system with limited interaction possibilities is capable of performing this type of observation task as effectively as direct manipulation, we performed a user study where we compared the participants' performance on different variants of trajectory modulation with regard to direct manipulation.

**Related work**

Teleoperation is a control mode regularly used for handling dangerous products such as radioactive products (Tokatli et al., 2021), or to reduce the cognitive and physical constraints of a task as in surgery (Santos Carreras, 2012). One of the main issues is to transcribe a user's movement in a secure and efficient way. The choice generally made is to let the user directly control the robot's movements, while managing the safety risks that may be caused by these movements. This can be done by communicating any risk of exceeding constraints to the user and prohibiting any command that could cause this situation. This communication can be visual via the display of a virtual double (Pan et al., 2021), or haptic (Lin et al., 2018). To ensure that the desired movement is achieved, the tool used can even be an arm identical to the arm used for manipulation (Singh et al., 2020). The main limitation of this solution is that it generally only communicates an impossibility to perform the task, and not a solution to achieve it. For large-amplitude movements in orientation, finding such a solution is often unintuitive, requiring relearning how to perform the task specifically through the platform (Sakr et al., 2018). Nevertheless, for relatively simple tasks which movements can be classified into a small number, it can be interesting to plan them in advance and let the user choose from among them. This approach is applied in particular in (Aleotti & Caselli, 2012), in which the user can intuitively choose a way of catching objects from a list of available methods planned in advance.

In addition, measuring the effectiveness of the proposed solution can only be done through performance criteria. The classic approach is to measure two types of indicators when the system is used by a group of users. On the one hand, objective criteria e.g. time taken to complete the task (Naceri et al., 2021), are recorded using measuring instruments. On the other hand, subjective criteria related to the quality of the interface, are measured using standardised questionnaires completed by users such as the SUS (Brooke, 1995) or the NASA TLX (Hart, 2006). As the criteria for evaluating the performance of an interface are highly dependent on the chosen task as the feeling of acceptability (Brooke, 1995) and their relevance can hardly be guaranteed a-priori such as the feeling of user agentivity (Sagheb et al., 2023), a set of criteria is often used for this evaluation.

**USER STUDY**

**Choice of Experimental Task**

The application task involves visually classifying dangerous products and physically handling them. Given that this task is dangerous and requires a high level of expertise, the size of the test cohort and the opportunities for testing on them are too small to be able to properly assess the performance of an interface. This is why we needed to use a task with similar characteristics to the application task, on which as many people as possible had expertise.

The task we chose was to read words on white capsules (dimensions 6 x 12 mm) placed into cylindrical vials (dimensions 16 mm x 70 mm). As with the application task, this task requires to observe visual features in a vial of a moving object (requiring slow, fine movements) for visual classification. In addition, it was likely that the majority of participants had a reading expertise; a significant difference between direct handling performance and performance via an interface could therefore be illustrated more easily.

## Platform Description

The platform was based on ROS middleware. As shown on Fig. 1, a Franka-Emika Panda robotic arm manipulated the vials in front of a Logitech Brio 4K monocular camera. The 3D pose of the vial in relation to the monocular camera was obtained using an Optitrack motion capture system.

The paths followed by the robot were pre-defined in relation to the position of the camera's optical centre, as being achievable whatever the robot's fixed capabilities (speed, acceleration), and as being necessary to perform the task. Two different paths could be selected, and were computed using the MoveIt software. Once a feasible trajectory was computed from one of those paths by an online trajectory generator, it was used as the input for the robot controller.

As illustrated in Fig 1, users interacted with the platform seated, in front of a workstation composed of a computer and two screens. Users were asked to adjust the elements of the workstation to ensure the best possible postural comfort (desk height, screens orientation). The first screen was used to observe the vials through different interfaces. The second screen was used to display a reminder of how to use the different proposed modalities.

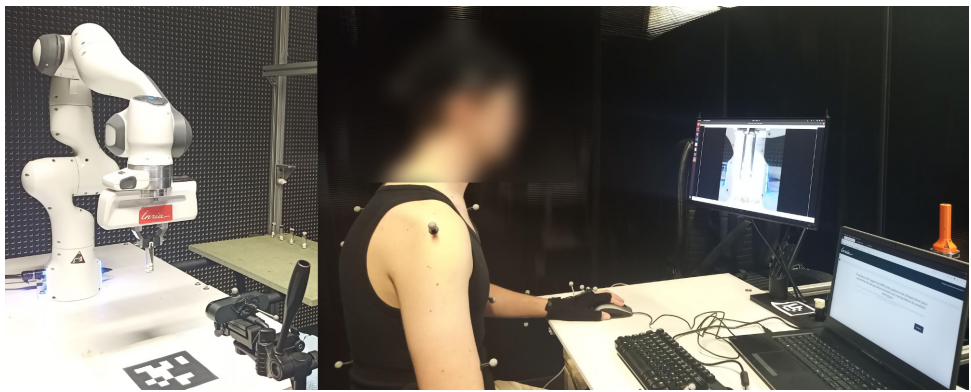

**Figure 1:** Representation of platform for vial manipulation (left) and user side (right).

## Study Design and Procedure

We therefore wanted to check whether a remote manipulation modality with trajectory modulation performed as well as the direct manipulation modality, and the contribution of the types of interaction proposed in the remote modalities. To check this, we carried out a within-subject study, in which subjects had to perform the reading task with each of the 5 modalities: one with direct manipulation, one under security conditions similar to the application task, and 3 via different remote handling methods. 37 participants took part in this experiment (age: 27±5, 20 females), approved by an ethics committee (Inria Coerle 313 AUCTUS). The conditions for inclusion were that the user was over 18 and affiliated to a social security scheme, had no chronic physical pathology and no major uncorrected visual problems. At the start of the experiment, participants were informed of the confidentiality of their personal data, received a presentation of the study and signed a consent form.

The first modality performed by the participants was direct reading (DIR variant); the participants took the vials by hand and red their contents. In the second modality, participants took and red the vials placed behind a protective glass (80 x 50 cm) (variant SEC). The glass was placed in such a way as to reproduce the constraints induced by the fume hoods used for the application task (height of the glass). These first two modalities enabled the participant to grasp the characteristics of the task (reaction of the capsules to movement in particular). Once these modalities had been performed, the participant was invited to evaluate them via a questionnaire.

Then, three modalities were tested, in random order: passive observation of videos (POV); active observation of videos (AOV); and observation of vials manipulated in real time on the platform (MANP). AOV and POV modalities consisted in observing videos of the movement of a vial manipulated by the platform, produced prior to the experiment. In these videos, the robot followed a minimal generic path to be able to read the contents of the vial, in front of the camera, at a fixed speed. The robot's path and trajectory were therefore identical whatever the video. These videos were selected randomly from a dataset of videos, and were observed using VLC software. In variant AOV, participants could interact with the video in a number of ways: pause it (then called "pauseB"), zoom on a particular point ("zoom"), speed up or slow down the video ("speedB"), move the video forward or backward by 10 seconds ("timeJump"). Those interactions were performed via the software's graphical interface (except for "zoom", provided by the Ubuntu OS). In variant POV, participants were asked to observe the video without interacting with it.

In MANP modality, a vial held by a robotic arm travelled along a series of pre-determined geometric paths in front of a camera. Participants were invited to read the contents of the vial solely through visual feedback from the camera. They could interact with the trajectory of the vial via a graphical interface, based on the Rviz software and divided into three parts (as displayed in Figure 2). The first part corresponded to the visual feedback from the camera, centred on the vial and adapted to the vial - camera distance. A second part displayed a digital twin of the robot's movements. The last part (top right) was the graphical interface for controlling the robot's trajectory. The available interactions were: speeding up or slowing down velocity as absolute value (called next "speedC"; values = 0 - 30% of maximal articular speed), stopping ("pauseC"), changing the direction of the current path ("inverseMvt"), change from one geometric path to another ("changeSide"), zoom on a particular point ("zoom").  The zoom function was provided by the Ubuntu OS.

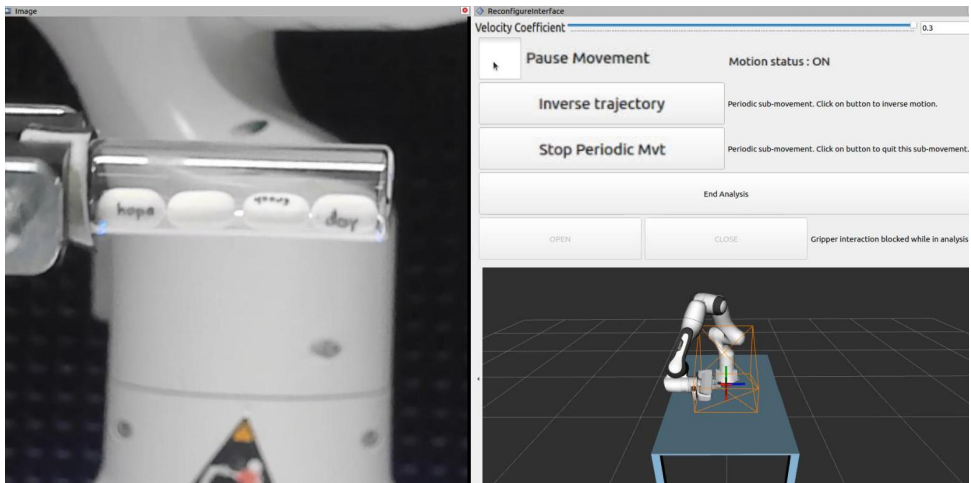

**Figure 2:** Representation of interface for MANP modality: visual feedback (left), digital twin (bottom right), interaction interface (top right). "Velocity coefficient", "Pause Movement", "Inverse trajectory", "Stop Periodic Mvt" corresponds to respectively "speedC", "pauseC", "inverseTraj", "changeSide".

Participants were given a quick introduction to the interfaces of variants AOV and MANP just before they started using these variants. Participants also completed a questionnaire after each of the POV, AOV and MANP variant. For each modality, 4 vials each containing 4 randomly selected capsules were red orally by the participant. Participants had a maximum of 120 seconds per vial to complete the task. This time corresponds to the duration of the robot's pre-programmed movement sequence in the AOV and POV modes. The transition from one vial to another took place when the participants felt they had completed their reading task or if a maximum observation time was exceeded.

### Data Collection

The following data were collected to assess each modality performances. The objective data recorded were the average time taken to complete the task over 4 vials (referred as "overall time"); if a perfect success was achieved during this modality ("perfect success"); and the number of each type of interaction performed for each of the vials in modalities AOV and MANP. A perfect success corresponded to a perfect reading of all the words in each of the 4 vials.

The subjective data recorded corresponded to the questionnaire completed by the participants after each of the modalities. This questionnaire, taking inspiration from (Naceri et al., 2021), particulary for evaluating the concepts of stress and discomfort, evaluated 4 characteristics of the interface. The **visual performance** of the interface was assessed as a mean of two notes: how easy it was to read the words in the vial and the visual feedback quality. The **temporal performance** of the interface was evaluated as a mean of the satisfaction of the time taken to complete the task and the speed of the robot. The **ease of use** of the interface was assessed as a mean of the ability to correctly predict the robot's behaviour, the absence of stress (defined as the level of anxiety felt when using the platform) and postural discomfort compared with direct handling, and the ease of completing the task. Finally, the **acceptability of the interface** was assessed using the standardised SUS questionnaire (Brooke, 1995). Apart from the acceptability of the interface (evaluated from 0 – 100), the questions were evaluated using a 5-point Likert scale (1: bad; 5: good).

Complementarily to the assessment of postural discomfort using a questionnaire, we assessed it using the RULA (McAtamney & Nigel Corlett, 1993) and REBA (Hignett & McAtamney, 2000) ergonomic grids. As the data we obtained confirmed that the postures adopted by the subjects in the different variants were too different for a comparison using ergonomic scores to be relevant, postural discomfort was only evaluated through a questionnaire.

### Statistical Analysis

Our first hypothesis was that one of the trajectory modulation modalities performed equally to the direct manipulation modality (Ha). To test this hypothesis, we compared the overall time, perfect success, and subjective scores obtained between all the modalities by all the subjects. In order to be able to interpret the overall time data (specifically, whether a short time could be associated with good performance), we assessed the correlation between overall time, perfect success and subjective scores.

Our second hypothesis was that the interactions proposed in the AOV and MANP modalities had a positive impact on the performance of these modalities (Hb). We tested this hypothesis by combining 3 criteria, ranked in ascending order of importance: the frequency of use of the interactions, the type of correlation between the frequency of use and the order of passage of the vials, and the type of correlation between the frequency of use and the other data recorded. The comparison of interaction frequencies illustrated whether an interaction was more or less appreciated, given that all the participants were well aware of all the interactions thanks to the training sessions prior to the trials. The correlation between frequency of use and order of use allowed us to identify whether the participants' increased experience led them to favour certain interactions over others. Finally, the level of correlation between frequency of use and the other data could be used to assess whether the use of the interactions leads to an improvement or deterioration in these criteria.

The Shapiro-Wilk test applied to the data rejected the normality hypothesis ($p < 0.05$). A Friedman's ANOVA (Field et al., 2012) was therefore applied to the data (except for the binary data "perfect success", on which a Cochran's Q test was applied). Significant tests were followed by a Wilcoxon post-hoc test with Bonferroni correction. For correlation tests, Spearman's correlation test was used.

### RESULTS

### Results about Objective and Subjective Data across Modalities

Analysis of the objective data showed that:
- There was no significant difference in the proportion of perfect successes between the variants ($\chi^2_{Cochran}(4) = 7.08$, $p = 0.13$); each variant had a majority proportion of perfect successes (0.756 for modalities POV and AOV, 0.567 for modalities MANP, DIR, SEC).

- "Overall time" differed significantly between modalities ($\chi^2_{Friedman}(4) = 118$, $p < 0.001$); particularly between the DIR/SEC modalities and the POV/AOV/MANP modalities (as illustrated in Figure 3).
- There was a negative correlation between "overall time" and "perfect success" ($R(35) = -0.19$, $p = 0.047$). Apart from satisfaction with robot speed (where no significant correlation was found), there was significant negative correlations between "overall time" and the various subjective scores ($-0.52 < R(35) < -0.34$; $p < 0.05$).

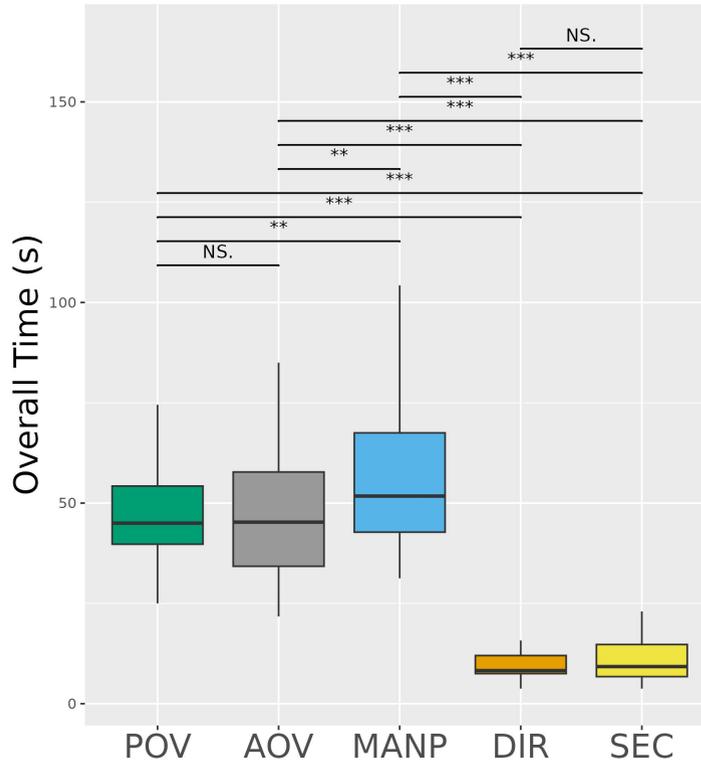

**Figure 3:** Mean time taken to complete the task, accross all variants. (Legend : NS. : $p \geq 0.05$; (*) : $p \leq 0.05$; (**) : $p \leq 0.01$; (***) : $p \leq 0.001$). Boxplot corresponds to median, upper / lower quartile, and scope of data.

Concerning the subjective data:
- The only criteria that differed significantly between the variants were: satisfaction with robot speed ($\chi^2_{Friedman}(2) = 10.5$, $p < 0.01$) for modalities AOV/MANP and POV ($p < 0.05$; medians: 4,4,3 ), and absence of postural discomfort ($\chi^2_{Friedman}(3) = 35.6$, $p < 0.001$) between modalities SEC and POV/AOV/MANP ($p < 0.001$; medians: 3,4,4,4 ).
- Among the POV/AOV/MANP variants, the median scores obtained for each of the interface evaluation criteria were respectively: 3.5/4/4 for temporal performance; 3.5/4/4 for visual performance; 4.25/4.25/4 for ease of use; and 85/85/80 for acceptability of the interface evaluated by the SUS score. The worst subjective score corresponded to the estimated ease of performing the task compared with direct handling (medians POV/AOV/MANP: 3/3/2).

## Results about Objective and Subjective Data across Interactions types of Modalities AOV and MANP

The 3 axes of analysis of the types of interaction for the AOV and MANP modalities are presented in tables 1 and 2.

**Table 1.** Analysis of the different types of interactions of modality AOV according to the proposed axes (NS. : not significant ; "subjective" corresponds here to ease of use, temporal and visual performance of the interface).

| Interaction name | Median interaction number (on 45 seconds) | Correlation with vial running order | Correlation with objective / subjective criteria |
|---|---|---|---|
| **pauseB** | 2.6 | NS. | NS. |
| **speedB** | 0.6 | NS. | NS. |
| **timeJump** | 0.35 | $R(146) = -0.18$; $p = 0.026$ | NS. |
| **zoom** | **0.0** | NS. | $-0.52 < R(35) < -0.34$; $p < 0.05$ (subjective) |

**Table 2.** Analysis of the different types of interactions of modality MANP according to the proposed axes (NS. : not significant ; "subjective" corresponds here to ease of use, temporal performance, and acceptability of the interface).

| Interaction name | Median interaction number (on 45 seconds) | Correlation with vial running order | Correlation with objective / subjective criteria |
|---|---|---|---|
| **pauseC** | 1.7 | NS. | $R(35) = 0.44$; $p = 0.007$ (overall time) |
| **inverseMvt** | 0.27 | $R(146) = -0.18$; $p = 0.028$ | $R(35) = 0.41$; $p = 0.012$ (overall time) |
| **changeSide** | 0.25 | NS. | $R(35) = 0.41$; $p = 0.012$ (overall time) |
| **speedC** | 0.12 | NS. | NS. |
| **zoom** | **0.0** | NS. | $-0.57 < R(35) < -0.34$; $p < 0.05$ (subjective) |

### DISCUSSION AND FUTURE WORK

These results showed that our task could indeed be performed using the interface. The proportion of successes was similar between all modalities, and high ratings of the various aspects of the platform were obtained via the subjective questionnaire (temporal, visual performance and ease of use > 3/5, acceptability above the acceptability threshold (70) defined by the SUS whatever the variants; less postural discomfort felt on the variants than for manipulation in secure conditions). However, our results also showed that the performance of the platform did not match that of direct handling. This can be found by the low scores for the estimation of the ease of performing the task compared with direct handling. This is also illustrated by the much longer times taken to complete the task on the POV/AOV/MANP variants compared with the DIR/SEC variants; given that a shorter time is correlated with a higher proportion of successes and better subjective scores, a shorter time may be associated with better performance. (Ha) hypothesis could not be validated.

The contribution of the interactions proposed for modalities AOV and MANP can also be questioned. These interactions didn't seem to be linked to better results, but even to a deterioration in these results. This can be found by the absence of any significant difference between the POV and AOV variants, whereas the main difference between these modalities is the presence of interactions. This is also illustrated by the combination of several factors concerning certain interaction: for example, the "zoom" interaction (variants AOV/MANP), which is not only used very little, but whose frequency of use is also proportional to a deterioration in subjective scores. Again, hypothesis (Hb) could not be validated. With regard to MANP mode, the tests also showed that the proposed interactions did not always compensate for the flaws in the interface. One of which was the limited number of geometric paths available, due to the difficulty of obtaining them (currently achieved empirically, to guarantee execution of the movement whatever the speed and acceleration of the robot). This was illustrated by the absence of a shaking movement to unblock capsules in vials; the different types of interaction available did not always allow this unblocking, which could prevent the observation task from being carried out. In our study, this resulted in the rejection of 7 participants and a final sample of 37 participants, given that this incident cannot happen in the application task.

These results therefore seem to suggest that directly manipulating the robot's movements could improve task execution performance. Using a conventional industrial robot resulted in too long execution times for such manipulation; it would be therefore necessary to use a serial manipulator designed for this task. Such a manipulator will be tested in a new experiment, on a more difficult task, to ensure that handling issues could be reflected in the ability to perform the task.

## CONCLUSION

We tested a new way of deferring the observation of a manipulated object, by only allowing the user to control the trajectory of a robotic arm holding this object. Our hypothesis was that this limited mode of control was sufficient to guarantee a safer manipulation but at least as effective as direct manipulation. We tested this hypothesis using a reading task inspired by an industrial chemical handling task. Analysis of the performance showed the proposed control limited to the trajectory was not as efficient as direct manipulation. An interface allowing more reactive manipulation of the vial's movements seems necessary, and will be tested in a future experiment.

## ACKNOWLEDGMENT

This project was funded by the French region "Nouvelle-Aquitaine" in the framework of the "Miels" project in collaboration with Solvay. We would like to thank Lucas Joseph for his help on the robotic experimental set-up, and Benjamin Camblor for his help on the study design.

**REFERENCES**

Abbott, Steven, and Charles M. Hansen (2008). Hansen solubility parameters in practice. Hansen-Solubility, ISBN 978-0-9551220-2-6

Aleotti, J., & Caselli, S. (2012). Grasp programming by demonstration in virtual reality with automatic environment reconstruction. Virtual Reality, 16(2), 87–104. https://doi.org/10.1007/s10055-010-0172-8

Berscheid, L., & Kröger, T. (2021). Jerk-limited Real-time Trajectory Generation with Arbitrary Target States (arXiv:2105.04830). arXiv. http://arxiv.org/abs/2105.04830

Brooke, J. (1995). SUS: A quick and dirty usability scale. *Usability Eval. Ind.*, *189*.

Field, A., Miles, J., & Field, Z. (2012). Discovering Statistics Using R. SAGE Publications Ltd

Hart, S. (2006). Nasa-task load index (Nasa-TLX); 20 years later. In *Proceedings of the Human Factors and Ergonomics Society Annual Meeting* (Vol. 50). https://doi.org/10.1177/154193120605000909

Hignett, S., & McAtamney, L. (2000). Rapid Entire Body Assessment (REBA). *Applied Ergonomics*, *31*(2), 201–205. https://doi.org/10.1016/S0003-6870(99)00039-3

Lin, Y., Zhao, H., Zhang, Y., & Ding, H. (2018). A Data-Driven Motion Mapping Method for Space Teleoperation of Kinematically Dissimilar Master/Slave Robots. 2018 IEEE International Conference on Robotics and Biomimetics (ROBIO), 862–867. https://doi.org/10.1109/ROBIO.2018.8665119

McAtamney, L., & Nigel Corlett, E. (1993). RULA: A survey method for the investigation of work-related upper limb disorders. *Applied Ergonomics*, *24*(2), 91–99. https://doi.org/10.1016/0003-6870(93)90080-s

Naceri, D., Mazzanti, D., Bimbo, J., Tefera, Y., Prattichizzo, D., Caldwell, D., Mattos, L., & Deshpande, N. (2021). The Vicarios Virtual Reality Interface for Remote Robotic Teleoperation: Teleporting for Intuitive Tele-manipulation. *Journal of Intelligent & Robotic Systems*, *101*. https://doi.org/10.1007/s10846-021-01311-7

Pan, Y., Chen, C., Li, D., Zhao, Z., & Hong, J. (2021). Augmented reality-based robot teleoperation system using RGB-D imaging and attitude teaching device. Robotics and Computer-Integrated Manufacturing, 71, 102167. https://doi.org/10.1016/j.rcim.2021.102167

Sakr, S., Daunizeau, T., Reversat, D., Regnier, S., & Haliyo, S. (2018). A Handheld Master Device for 3D Remote Micro-Manipulation. 2018 International Conference on Manipulation, Automation and Robotics at Small Scales (MARSS), 1–6. https://doi.org/10.1109/MARSS.2018.8481194

Sagheb, S., Gandhi, S., & Losey, D. P. (2023). Should Collaborative Robots be Transparent? (arXiv:2304.11753). arXiv. https://doi.org/10.48550/arXiv.2304.11753

Santos Carreras, L. (Ed.). (2012). Increasing Haptic Fidelity and Ergonomics in Teleoperated Surgery. EPFL. https://doi.org/10.5075/epfl-thesis-5412

Singh, J., Srinivasan, A. R., Neumann, G., & Kucukyilmaz, A. (2020). Haptic-Guided Teleoperation of a 7-DoF Collaborative Robot Arm With an Identical Twin Master. IEEE Transactions on Haptics, 13(1), 246–252. https://doi.org/10.1109/TOH.2020.2971485

Tokatli, O., Das, P., Nath, R., Pangione, L., Altobelli, A., Burroughes, G., Jonasson, E. T., Turner, M. F., & Skilton, R. (2021). Robot-Assisted Glovebox Teleoperation for Nuclear Industry. Robotics, 10(3), Article 3. https://doi.org/10.3390/robotics10030085